\documentclass[]{article}
\usepackage{setspace}
\usepackage{amsmath}
\usepackage{graphicx}
\usepackage{subcaption}
\usepackage{xcolor}

\makeatletter
\newcommand*{\centerfloat}{%
  \parindent \z@
  \leftskip \z@ \@plus 1fil \@minus \textwidth
  \rightskip\leftskip
  \parfillskip \z@skip}
\makeatother

\begin{document}
%
\title{Extending Defensive Distillation}
\author{Nicolas Papernot and Patrick McDaniel\\
	Pennsylvania State University\\
		\{ngp5056,mcdaniel\}@cse.psu.edu
}
\date{}


\maketitle

\begin{abstract}
Machine learning is vulnerable to adversarial examples: inputs carefully
modified to force misclassification. Designing defenses against such inputs remains largely an
open problem. In this work, we revisit defensive distillation---which is one of the
mechanisms proposed to mitigate adversarial examples---to address its
limitations. We view our results not only as an effective way of addressing some
of the recently discovered attacks but also as reinforcing the importance of
improved training techniques.
\end{abstract}

\vspace*{-0.1in}

%

\section{Introduction}
\label{sec:introduction}

\vspace*{-0.12in}

Deployed machine learning (ML) models are vulnerable to inputs maliciously 
perturbed to force them to mispredict~\cite{lowd2005adversarial,barreno2006can}. A class of such inputs, 
named \emph{adversarial examples}, are systematically constructed through slight
perturbations of otherwise correctly classified 
inputs~\cite{szegedy2013intriguing, biggio2013evasion}. These 
perturbations are chosen to maximize the model's prediction error
while leaving the semantics of the input
unchanged.
Although this often poses a non-tractable optimization problem for popular
architectures like deep neural networks, heuristics allow
the adversary to find effective perturbations---typically through the evaluation of gradients of
the model's output with respect to its inputs~\cite{szegedy2013intriguing,goodfellow2014explaining}.

To defend against adversarial examples, two classes of approaches exist. The first algorithmically improves upon the learning
to make the model inherently more
robust: techniques that fall in this class include adversarial training~\cite{szegedy2013intriguing,goodfellow2014explaining}
or defensive distillation~\cite{papernot2016distillation}.
The second is a set of detection mechanisms used to reject inputs 
suspected to be malicious. One approach is to analyze internal model features for anomalies~\cite{metzen2017detecting,feinman2017detecting}.
Another approach chooses to add an outlier 
class to the set of existing outputs that make up the ML task, and 
then train the model to map adversarial examples
to this special class~\cite{grosse2017statistical,hosseini2017blocking}.

However, most---if not all---of these defenses fail to adapt to novel attack strategies.
Defensive distillation, which is the primary mechanism discussed in the present document, is no exception. It is successful
against attacks known at the time of writing, such as
the Fast Gradient Sign Method (FGSM)~\cite{goodfellow2014explaining}
and the Jacobian-based Saliency Map Approach (JSMA)~\cite{papernot2016limitations}.
However, as advancements found new ways to mount attacks against
ML, defensive distillation
can now be evaded~\cite{papernot2017practical,carlini2016towards}.

One of the successful strategies is to mount a black-box attack, as 
shown in~\cite{papernot2017practical}: the adversary first trains an
\emph{undefended} surrogate 
model which mimicks the defended model, and then uses the surrogate model to generate adversarial examples that transfer back to the distilled model (i.e., the same inputs are misclassified by both the undefended and defended models). This attack strategy succeeds because
of the often strong transferability of adversarial examples across models trained
to solve the same ML task~\cite{szegedy2013intriguing,goodfellow2014explaining,tramer2017space}

In addition, distillation can be evaded with optimization attacks
\`a la Szegedy et al.~\cite{szegedy2013intriguing}, which have
been revisited in~\cite{carlini2016towards} with new objectives
and optimizers.
This second attack strategy is successful against defensive distillation
because of a phenomenon called \emph{gradient masking}~\cite{papernot2017practical}: 
the defense mechanism destroys
gradients essential to the heuristics of attacks like the FGSM and JSMA---instead of reducing the model's error. Note that defensive distillation is not the only mechanism that yields to gradient masking~\cite{brendel2017comment}


Given the two failure modes identified above, we propose in this work 
a variant of defensive distillation that addresses them. 
We demonstrate that our approach
is less susceptible to transferability and gradient masking by mounting black-box attacks from both undefended and defended surrogate
models. \emph{Like the original defensive distillation,
the technique does not require that the defender
generate adversarial examples.
Its applicability is thus less likely to be limited to
specific adversarial example heuristics.}

Unfortunately, it is currently infeasible to formally prove
 robustness guarantees for models like deep neural networks.
Techniques for the verification of these models are also still in their infancy~\cite{huang2016safety,katz2017reluplex}.
We thus resort to experimental validation of our approach and leave a formal
analysis to future work.

\section{Defensive distillation}
\label{sec:background}

\vspace*{-0.1in}

Adapted from~\cite{hinton2015distilling}, defensive distillation successively trains two instances of the same deep neural network architecture.\footnote{As noted in~\cite{papernot2016distillation}, distillation is applicable to any model outputting an energy-based probability distribution for which a temperature can be defined.} The
first model is trained with the original dataset $\{(x,y)\}$ where $y$
indicates the \emph{expected} correct class for the input $x$. Learning is performed conventionally to
the exception of the softmax\footnote{The term softmax refers here  to the last layer of a neural network classifier. It transforms scores assigned by the model to each class of the task into probabilities.} temperature, which is usually set to $1$ and is
here increased. In the interest of space, we refer
readers to the detailed presentation in~\cite{papernot2016distillation}.
Briefly, the model outputs probabilities that are closer
to a uniform distribution at high softmax temperatures.
This first model $f$ is then used to label each training point $x$ with its probability vector (roughly indicating how likely $f$ believes the input $x$ to be in each class of the task).
This defines a newly labeled training set $\{(x, f(x))\}$, which
the second model $f^d$ is trained on.
When deployed with a temperature of $1$, the model $f^d$ is found to be
robust to the FGSM and JSMA attacks.

\section{Extending defensive distillation}
\label{sec:approach}

\vspace*{-0.1in}

\begin{figure}[t]
	\centering
	\includegraphics[width=\textwidth]{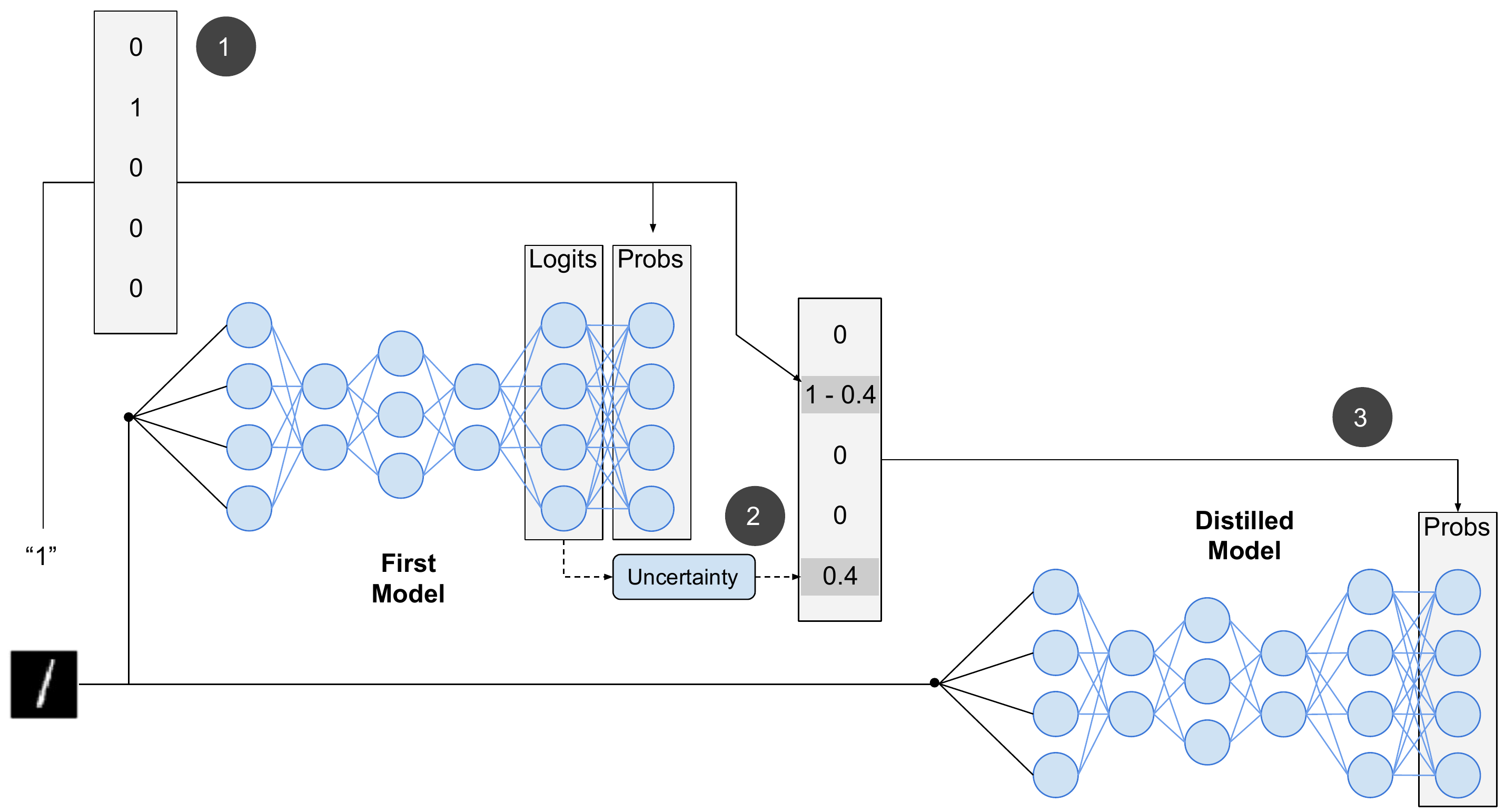}
	\caption{The extended defensive distillation procedure: (1) the first neural network
	is trained as usual on one-hot labels, (2) a new labeling vector is defined for each training point by combining
	the original label information together with the first model's predictive uncertainty---which is measured
	by taking several stochastic passes through the model to infer the logits,
	(3) the distilled model is trained at temperature $T\geq1$
	with the new label vectors.}
	\label{fig:extending-distillation}
\end{figure}

We revisit the defensive distillation approach by modifying the
labeling information used to train the distilled model $f^d$. 
Our changes are motivated by two lines of previous work. The first
augments the output of models with an outlier class as a means to mitigate adversarial
examples~\cite{grosse2017statistical,hosseini2017blocking}: the model
learns to classify adversarial inputs in the outlier class. The second 
provides uncertainty estimates in neural networks through stochastic inference~\cite{gal2015dropout,gal2016uncertainty}. 

The variant of defensive distillation described below successively trains
two models and transfers the knowledge of the first model $f$ into the second.
The mechanism enables the second (i.e., distilled) model $f^d$ to be uncertain 
about its prediction when the input it is asked to
label is far from the training manifold.
Uncertainty estimates required to train this distilled model are obtained through an
analysis of the predictions made by the
first of the two models.

\paragraph{Measuring uncertainty} To capture the uncertainty of the first
model in the labels created to train the second---distilled---model, we use dropout
at inference. Dropout is a technique which randomly ``drops'' (i.e., removes) some
of the neurons in the neural network's architecture. It is typically applied during training as a
regularizer. Instead, Gal and Ghahramani apply dropout while inferring, in order to approximate Bayesian inference~\cite{gal2015dropout}.
We use here a similar process to quantify the predictive uncertainty of the first model $f$.

While dropout is still activated, we take $N$ forward passes through the neural
network $f$ and record the $N$ logit vectors $z^0(x), ..., z^{N-1}(x)$ predicted.
Each logit vector $z^m(x)$ is $n-1$ dimensional, where $n$ is the number
of classes in the problem. The logits are
scores assigned to each class, before they are transformed into a probability 
through the application of a softmax at the output of the network.
For instance, $z^m_j(x)$ is the score assigned by model $f$ to class $j$ in the
$m$-th forward pass on input $x$ with dropout.

We then compute the mean logit vector $\bar{z}(x)$ and the variance over the $N$
stochastic passes to
measure uncertainty at the input point $x$ considered:

\begin{equation}
\label{eq:uncertainty-score}
\sigma(x) = \frac{1}{N} \sum_{m\in 0 .. N-1} \left( \sum_{j\in 0..n-1} \left(z^m_j(x) - \bar{z}_j\right)^2 \right)
\end{equation}

Unlike what was proposed in the original defensive distillation
mechanism, the softmax
temperature of the first model $f$ is maintained at $T=1$ at all times.

\paragraph{Training the distilled model with uncertainty} The output of the distilled model
is augmented with an outlier class, which is separate from the existing $n$ ``real'' classes
of the problem. When training the distilled model $f^d$, each training point is relabeled
with a vector that indicates (1) the correct class of the input and (2)
the uncertainty $\sigma(x)$ of the first model $f$. The 
components $j\in 0 ..n$  of this new labeling vector $k(x)$
are defined by:
\begin{equation}
\label{eq:revisited-distillation}
k_j(x) = \left\{
			\begin{array}{ll}
			1-\alpha \cdot \frac{\sigma(x)}{\max_{x\in \mathcal{X}} \sigma(x)} & \text{\ if\ } j = l \text{\ (correct class) }\\
			\alpha \cdot \frac{\sigma(x)}{\max_{x\in \mathcal{X}} \sigma(x)} & \text{\ if\ } j = n \text{\ (outlier class)}\\
			0 & \text{ otherwise}
			\end{array}
		\right.
\end{equation}
Informally, the uncertainty defines the probability assigned to the
outlier class, while the remaining probability is assigned exclusively to the correct class.
The parameter $\alpha$ weights the importance given to
the uncertainty measure. 
When the first model $f$ was uncertain on $x$, the measure $\sigma(x)$
is large and $k_n(x)$ as well. Hence, the distilled model is more likely
to classify uncertain inputs as outliers.
As a consequence, if $\alpha$ is set to a too large value,
all inputs are classified as outliers.
The distilled model $f^d$ trained with labeling vector $k(x)$ 
learns how to correctly classify $x$ as long as $k_l(x) > k_n(x)$ where
$l$ is the index of the correct label for $x$ and $n$ the index
of the outlier class.

Training with an outlier class is expected to increase the model's robustness
to finite perturbations (e.g., FGM
and JSMA adversarial examples), as observed by previous efforts including~\cite{grosse2017statistical,hosseini2017blocking}. We now make one
last modification to the training procedure in order to address smaller
perturbations (such as the ones produced by the AdaDelta strategy).
Specifically, we augment the training loss of our distilled model to include
the following penalty:
\begin{equation}
-\gamma \cdot \max\left(\max_{j\neq l, n} z_j(x) - z_l(x), \kappa\right)
\end{equation}
where $l$ is the correct class, $\gamma$ weights the relative contribution of the penalty in 
the training loss and $\kappa$ prevents unnecessary extrema during optimization.

Finally, we note that unlike in the original defensive distillation mechanism, we maintain
the temperature of the distilled model at all times equal to $T=1$. 

\section{Evaluation}
\label{sec:evaluation}

\vspace*{-0.1in}

We evaluate the robustness of the distilled model $f^d$ in the face of attacks mounted
both in \emph{white-box} and \emph{black-box} threat models. 
In addition to corresponding to a more realistic threat model,
robustness to black-box attacks is a compelling indicator of the
absence of gradient masking.
Below, the
strengths and weaknesses of the approach are characterized by four rates:
\begin{itemize}
	\item \emph{Accuracy}: percentage of legitimate inputs correctly classified.
	\vspace*{-0.1in}
	\item \emph{False positives}: percentage of legitimate inputs classified as outliers
	\vspace*{-0.1in}
	\item \emph{Recovered}: percentage of adversarial inputs whose
	class was recovered (i.e., they are assigned the label of the originally unperturbed input).
	\vspace*{-0.1in}
	\item \emph{Detected}: percentage of adversarial examples classified
	in the outlier class..
\end{itemize}
Ideally, the distilled model's accuracy should be comparable to the one
of an undefended model. Combined, the recovered and detected adversarial examples
should leave as few misclassified adversarial examples as possible. Finally,
the false positive rate should be as low as possible,
otherwise the model becomes unusable when presented with legitimate inputs.

When evaluating attacks, we also report the mean perturbation of adversarial
examples that were successful for the adversary (i.e., neither recovered or detected). This allows us to better visualize the effectiveness of the defense,
because it exposes the trade-off between indistinguishibility of adversarial examples and their likelihood of being misclassified.

\vspace*{-0.05in}

\subsection{Experimental setup}
\label{ssec:setup}

\vspace*{-0.05in}

We experiment with MNIST~\cite{lecun1998mnist} using
the convolutional neural network provided in the cleverhans v.1 library tutorials. 
Dropout layers are inserted before the input layer and after the last
convolution. Hence, dropout is applied to the feature representation
extracted by the convolutional layers, before the fully connected layer
essential to classification. The probability of dropping a neuron is set
to $0.2$ before the input layer and $0.5$ after the convolution.

We consider three attack strategies: the fast gradient 
method (FGM)~\cite{goodfellow2014explaining},
the Jacobian Saliency Map Approach (JSMA)~\cite{papernot2016limitations}
and the AdaDelta optimization strategy (AdaDelta)~\cite{carlini2016towards}.
For the JSMA and AdaDelta attacks, we randomly select the target class
among the set of ``real'' classes (i.e., we omit the outlier class since
it is not a profitable output for adversaries).
Adversarial examples are clipped to ensure their components are
within the range of acceptable values.

In our evaluation, the JSMA and AdaDelta attacks are ran on less test inputs than the FGM---they are computationally expensive. Thus, the variability of the results for these two attacks is larger than for results relative to the FGM.

Our source code will be open-sourced through cleverhans.

\subsection{Robustness to white-box attacks}

In the white-box setting, the adversary has access to the defended model's architecture
and parameters: the attack algorithms are directly applied to the model that is targeted (i.e., the
distilled model).

The distilled model was trained for $10$ epochs with $N=20$ dropout passes, a variance coefficient $\alpha=0.9$, a loss penalty weight $\gamma=5\cdot 10^{-4}$, and a loss penalty constant $\kappa=40$. 
This choice of hyper-parameters is justified below, except for $\gamma$ and $\kappa$: they were chosen with a grid search. The model accuracy on legitimate
test inputs is $97.28\%$, compared to a $98.41\%$ accuracy for the same architecture trained
without defensive distillation for $10$ epochs. The distilled model's false positive
remains below $1\%$.

\begin{figure*}[p]
	\centerfloat
	\begin{subfigure}[t]{0.65\textwidth}
		\includegraphics[width=\textwidth]{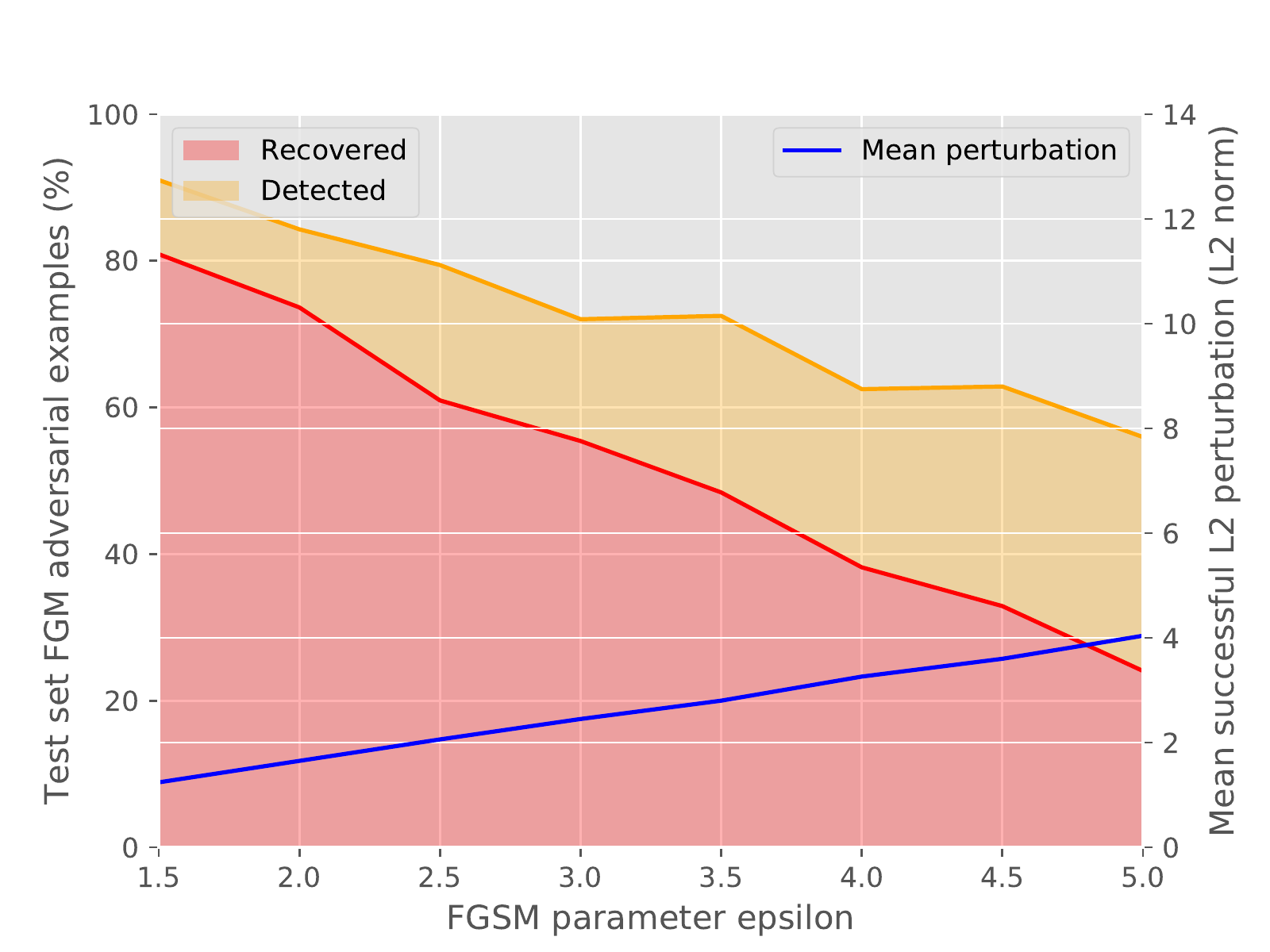}
		\caption{Robustness to FGM}
		\label{fig:white-box-fgm}
	\end{subfigure}%
	~\hspace*{0.1in}
	\begin{subfigure}[t]{0.65\textwidth}
		\includegraphics[width=\textwidth]{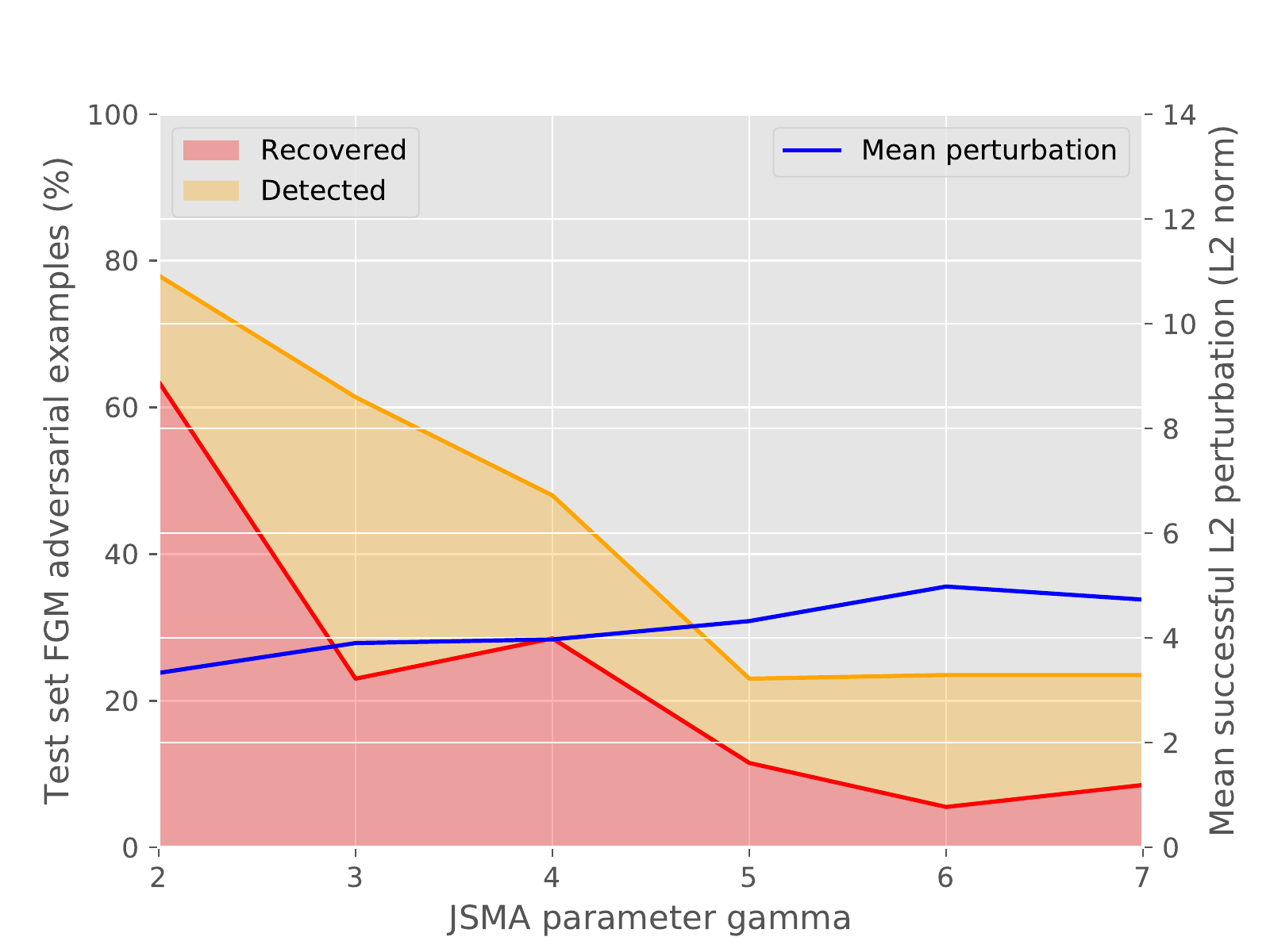}
		\caption{Robustness to JSMA}
		\label{fig:white-box-jsma}
	\end{subfigure}%
	\\
	\begin{subfigure}[t]{0.65\textwidth}
		\includegraphics[width=\textwidth]{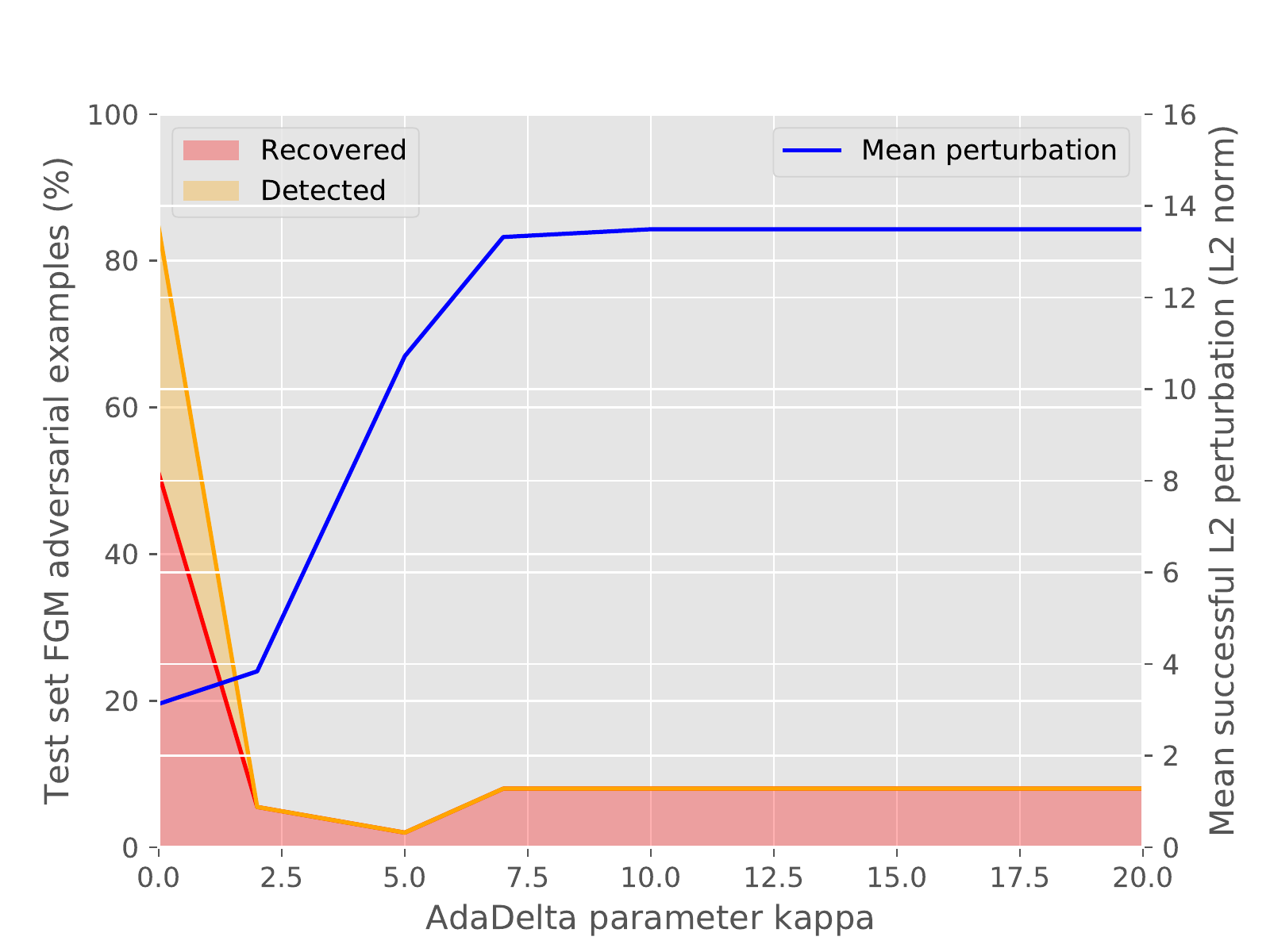}
		\caption{Robustness to AdaDelta}
		\label{fig:white-box-adadelta}
	\end{subfigure}%
	\\
	\begin{subfigure}[t]{0.65\textwidth}
		\includegraphics[width=\textwidth]{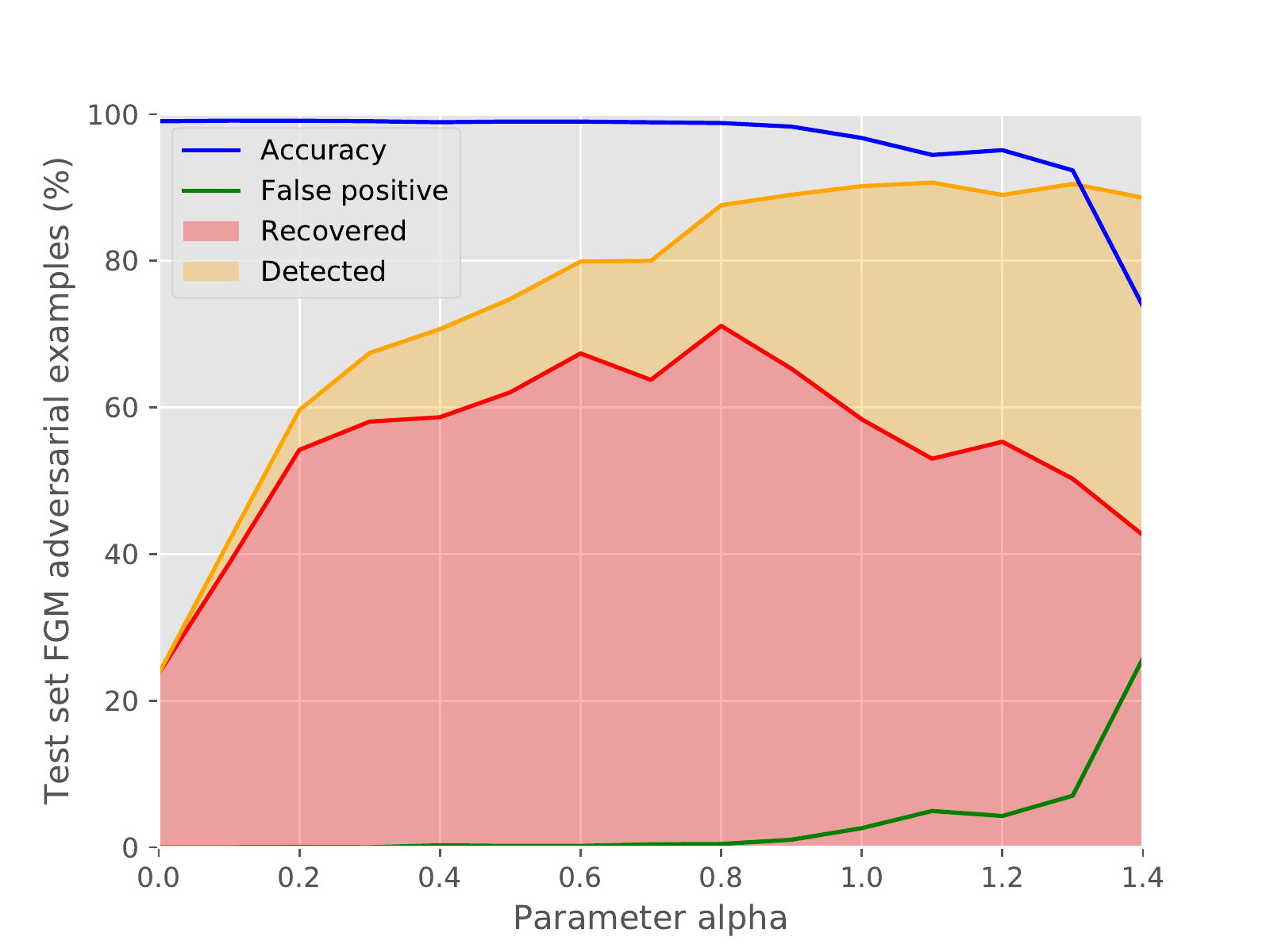}
		\caption{Exploration of parameter $\alpha$.}
		\label{fig:white-box-alpha}
	\end{subfigure}%
	~\hspace*{0.1in}
	\begin{subfigure}[t]{0.65\textwidth}
		\includegraphics[width=\textwidth]{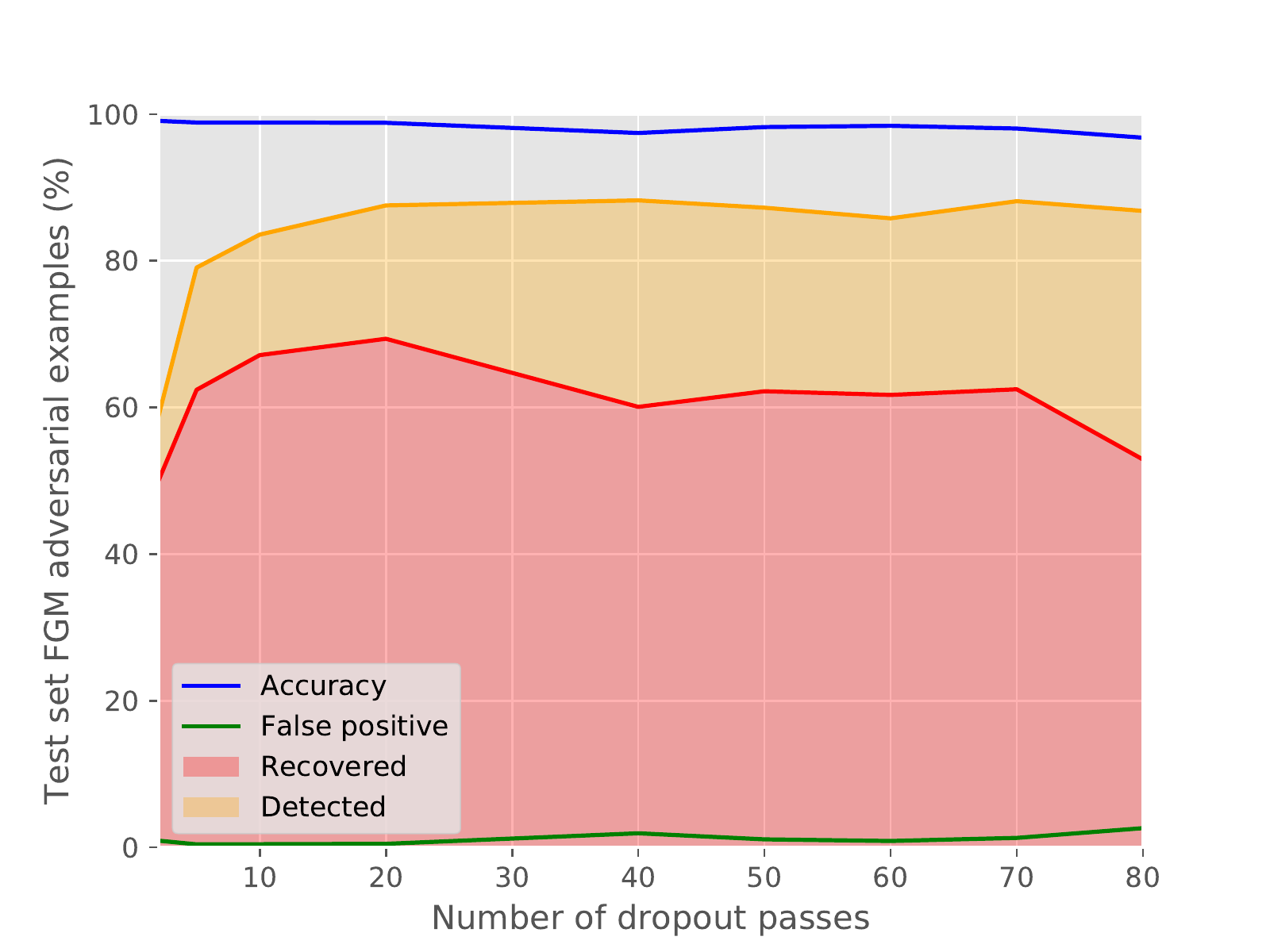}
		\caption{Exploration of the number $N$ of stochastic passes.}
		\label{fig:white-box-passes}
	\end{subfigure}%
	\caption{Evaluation of defensive distillation in the white-box setting. Figure~\ref{fig:white-box-fgm},~\ref{fig:white-box-jsma} and~\ref{fig:white-box-adadelta} report the recovered and detected rates vs. the mean perturbation of successful adversarial example produced with the three attacks 
		from Section~\ref{ssec:setup}. For clarity of presentation, the parameter exploration performed in Figures~\ref{fig:white-box-alpha} and~\ref{fig:white-box-passes}
		is plotted with the FGM only.}
	\label{fig:white-box}
\end{figure*}

\paragraph{Robustness}
We run the attacks mentioned in Section~\ref{ssec:setup} with different
parameters to evaluate the percentage of misclassified adversarial
examples (i.e., those that are neither recovered or detected) with
respect to their mean perturbation. Figure~\ref{fig:white-box-fgm},~\ref{fig:white-box-jsma} and~\ref{fig:white-box-adadelta} 
illustrate the results respectively for the FGM, JSMA and AdaDelta attacks.
We break down correctly classified adversarial examples as recovered
or detected, but stack the two rates to better visualize the inputs that
were correctly processed by the model.

The defended model is robust in a certain neighborhood of the test points.
In comparison, the undefended model has a misclassification rate of $90.8\%$, $92.2\%$ and $96.0\%$ on the
FGM, JSMA and AdaDelta attacks respectively.
As the attacks increase the perturbation introduced, the model
is less and less able to cope with adversarial input.
For the FGM and JSMA attacks, the mechanism is able to correctly classify a large fraction
of adversarial examples produced with a perturbation norm less than $4$ according to the $\ell_2$ norm.
For the AdaDelta strategy, there remains an evasion window for the adversary
using $2\leq\kappa\leq 2.5$.

A visual inspection reveals that when the $\ell_2$ norm of adversarial examples is
larger than $5$, the perturbations begin to modify 
the correct class (e.g., the digit corresponding to the wrong class predicted by the model appears 
transparently in the image or the digit corresponding to the correct
class is partially erased).

\paragraph{Parameter exploration} The fraction of adversarial examples
whose correct class is recovered or detected increases with the
value of $\alpha$, as illustrated in Figure~\ref{fig:white-box-alpha}. 
Indeed, larger values of $\alpha$ assign a greater probability to the outlier
class when creating the labeling vector $k(x)$ with which the distilled
model is trained.
However for $\alpha\geq 1.0$,
this comes at the expense of a lower accuracy on legitimate inputs: there are more
false positives---because legitimate inputs are increasingly likely
to be labeled by the model as outliers.

As more forward passes  with dropout are taken to estimate the uncertainty
of the first model, the distilled model is able to detect and recover more adversarial
examples. Yet, measurements depicted in Figure~\ref{fig:white-box-passes}  show that there is  a number of passes ($\simeq 20$) after which the curves exhibit a plateau. The approximation performed when taking successive stochastic passes eventually converges as the number of passes increases.

\subsection{Robustness to black-box attacks}

We mount worst-case black-box attacks through transferability from
a surrogate model trained using the same architecture and data.
For each attack, we consider two types of surrogate models: undefended ones (denoted by U) and 
ones (denoted by D) defended with the mechanism introduced in Section~\ref{sec:approach}.
This allows us to test for gradient masking (which would be indicated by  strong
transferability of adversarial examples from the surrogates to
the distilled model). 

\begin{figure*}[p]
	\centerfloat
	\begin{subfigure}[t]{0.65\textwidth}
		\includegraphics[width=\textwidth]{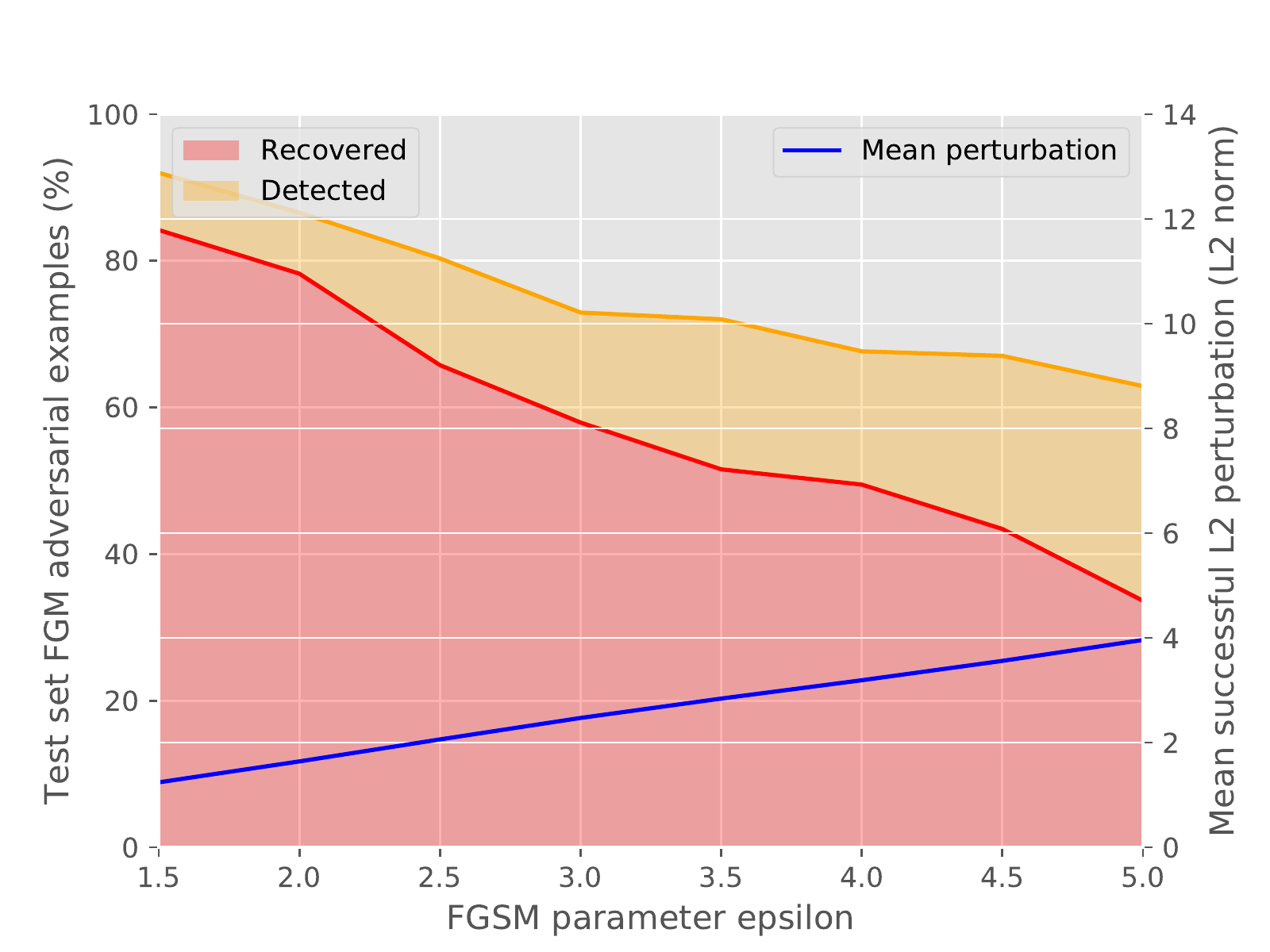}
		\caption{Robustness to defended FGM}
		\label{fig:black-box-fgm-defended}
	\end{subfigure}%
	~\hspace*{0.1in}
	\begin{subfigure}[t]{0.65\textwidth}
		\includegraphics[width=\textwidth]{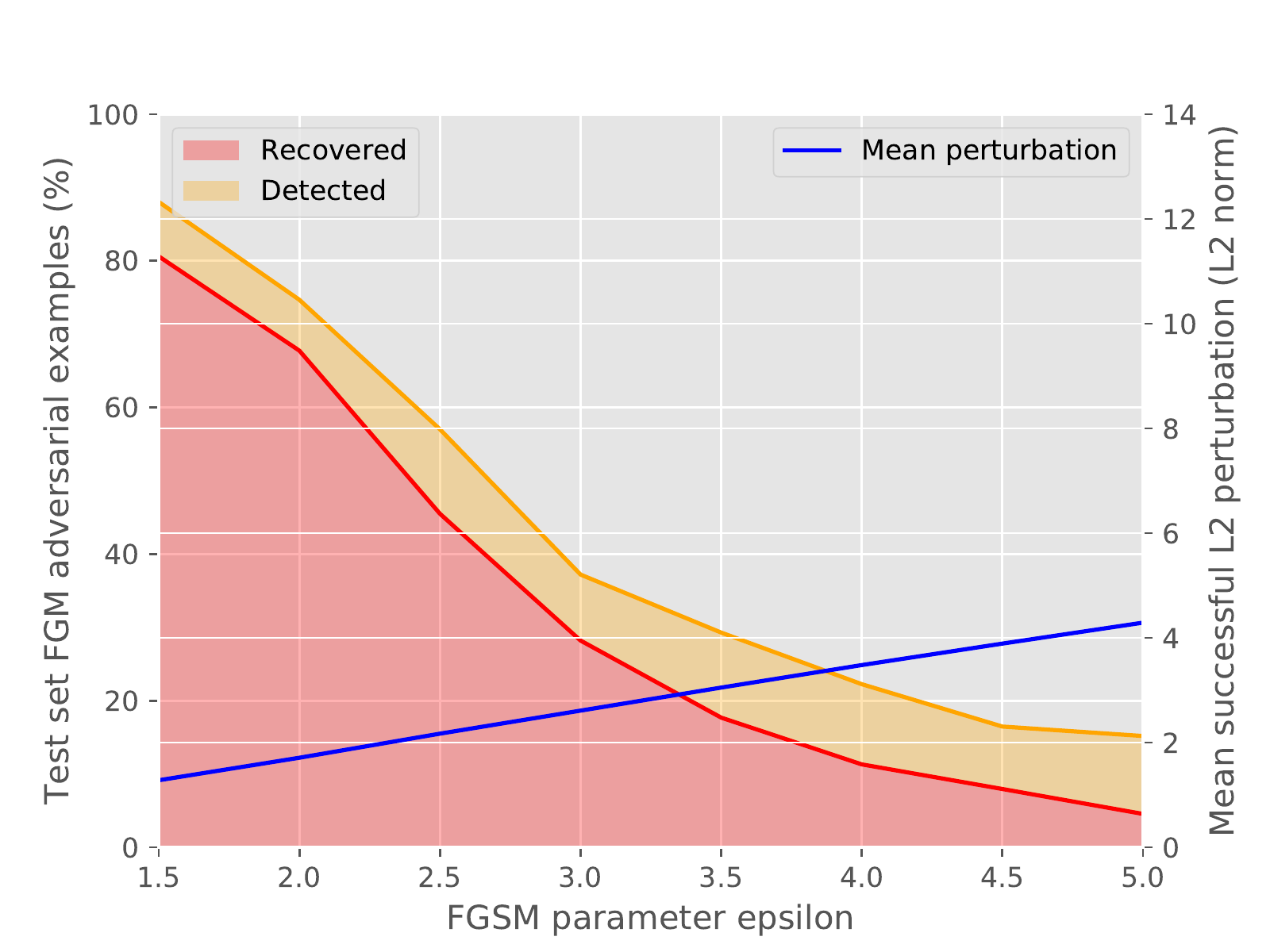}
		\caption{Robustness to undefended FGM.}
		\label{fig:black-box-fgm-undefended}
	\end{subfigure}%
	\\
	\begin{subfigure}[t]{0.65\textwidth}
		\includegraphics[width=\textwidth]{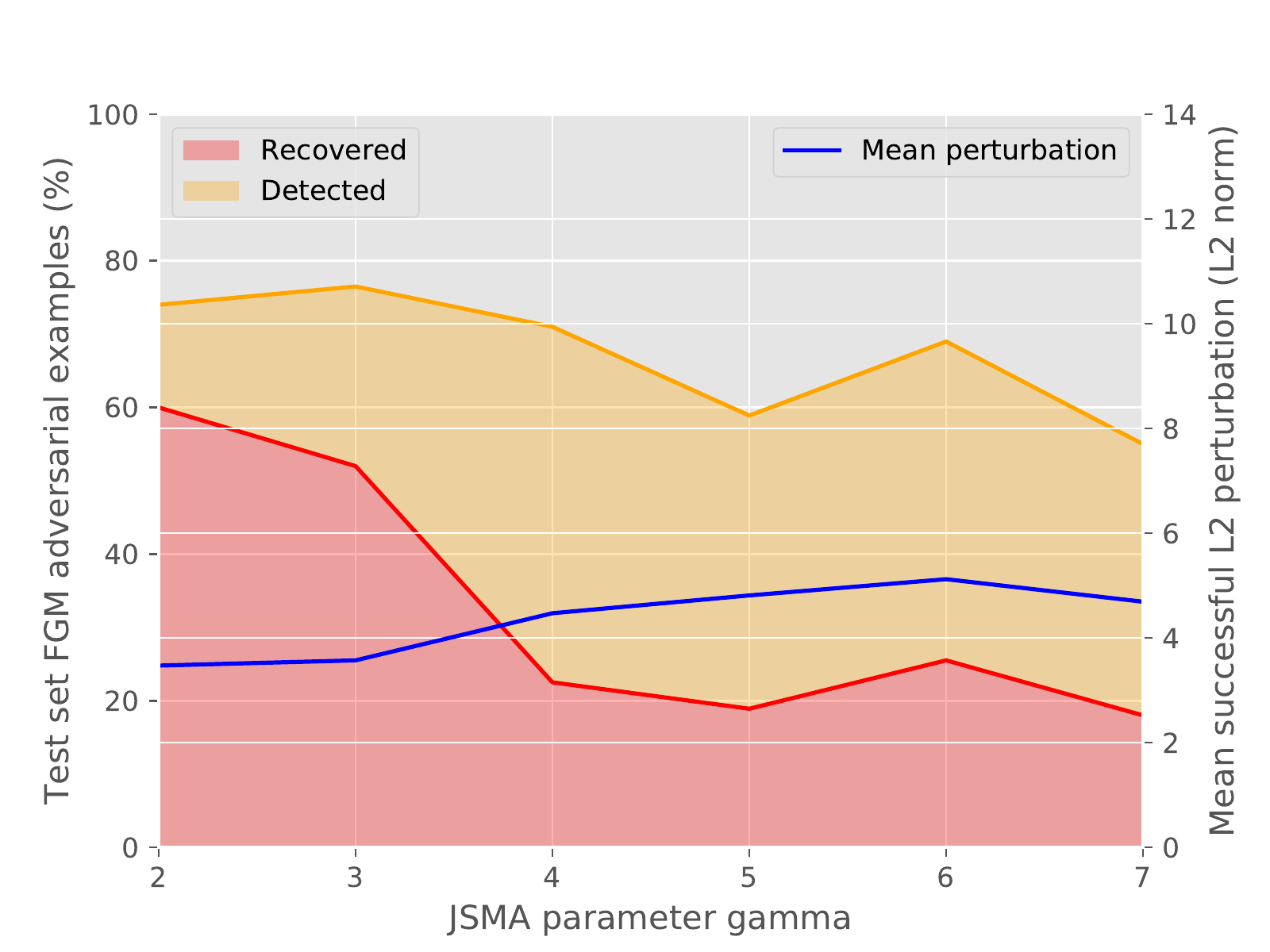}
		\caption{Robustness to defended JSMA}
		\label{fig:black-box-jsma-defended}
	\end{subfigure}%
	~\hspace*{0.1in}
	\begin{subfigure}[t]{0.65\textwidth}
		\includegraphics[width=\textwidth]{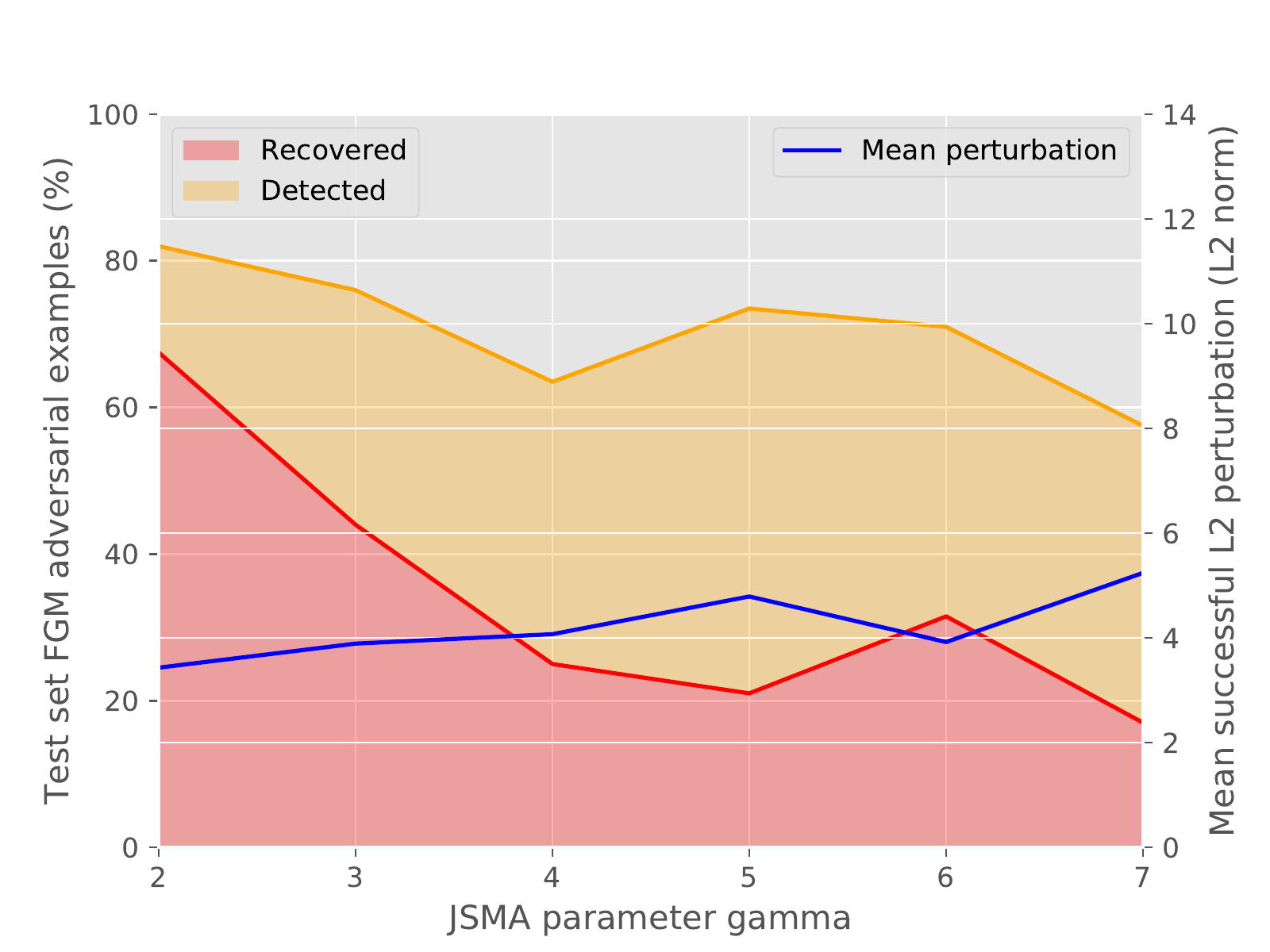}
		\caption{Robustness to undefended JSMA.}
		\label{fig:black-box-jsma-undefended}
	\end{subfigure}%
	\\
	\begin{subfigure}[t]{0.65\textwidth}
		\includegraphics[width=\textwidth]{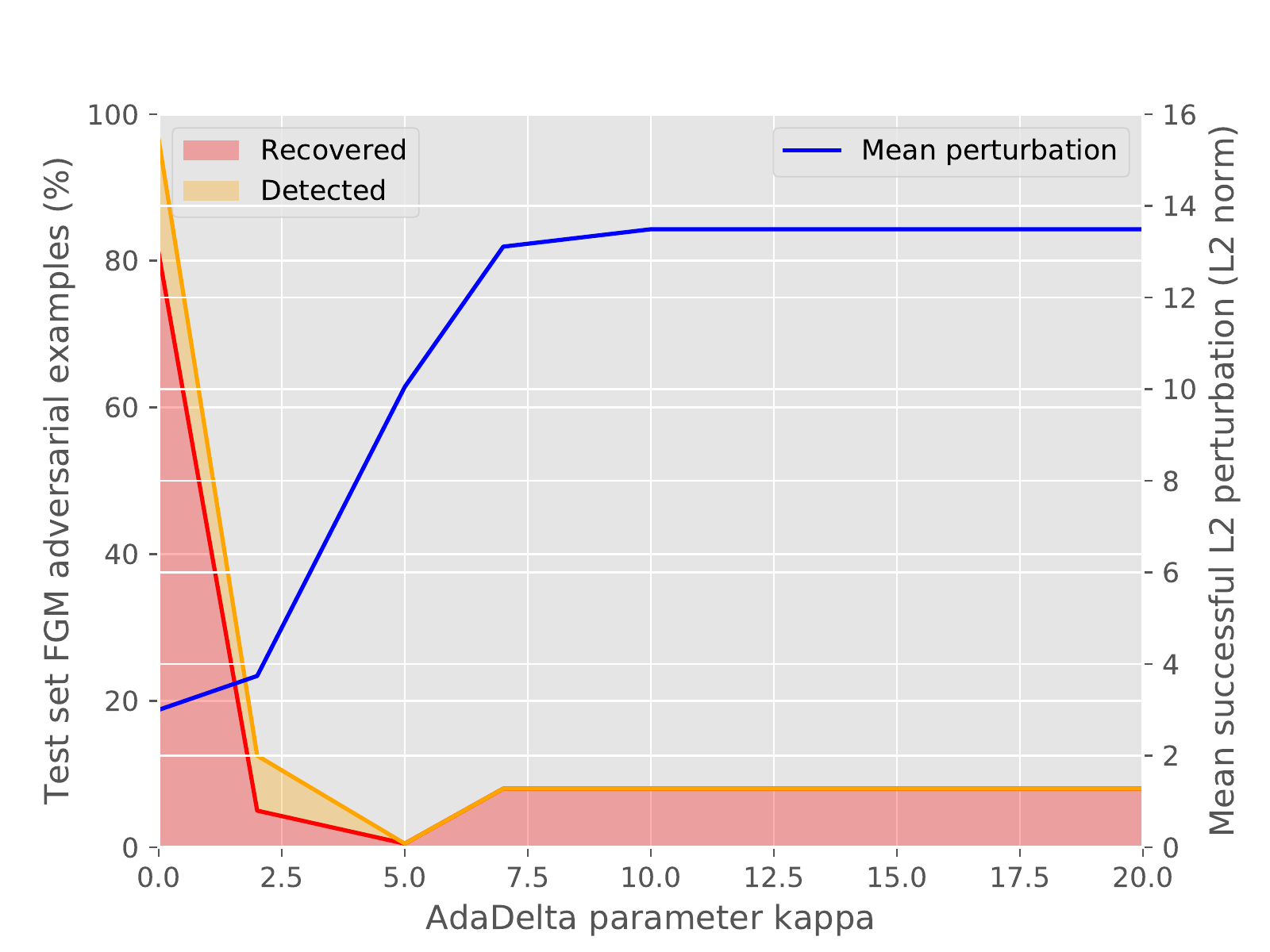}
		\caption{Robustness to defended AdaDelta}
		\label{fig:black-box-adadelta-defended}
	\end{subfigure}%
	~\hspace*{0.1in}
	\begin{subfigure}[t]{0.65\textwidth}
		\includegraphics[width=\textwidth]{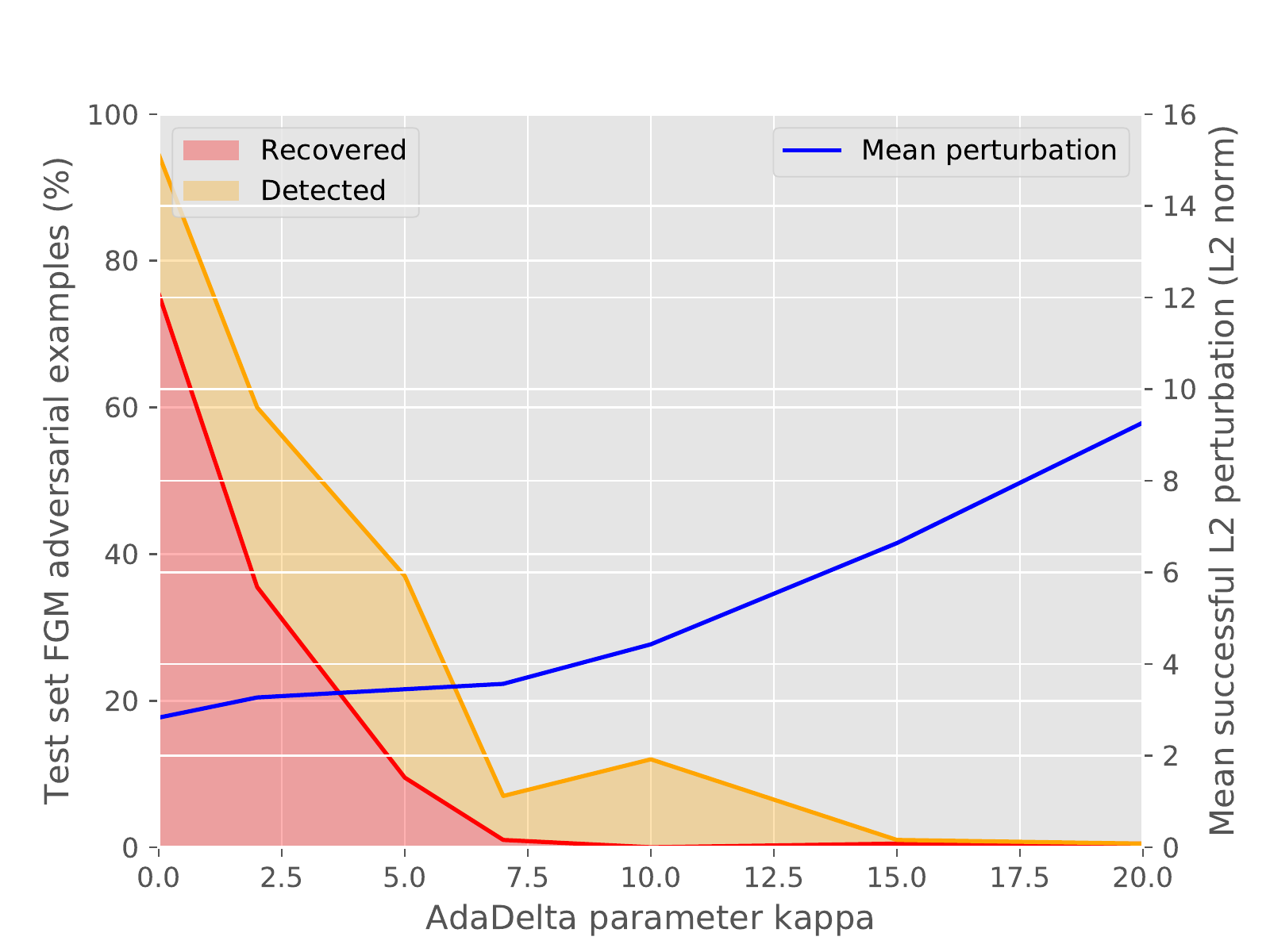}
		\caption{Robustness to undefended AdaDelta.}
		\label{fig:black-box-adadelta-undefended}
	\end{subfigure}%
	\caption{Evaluation of defensive distillation in the black-box setting. In Figures~\ref{fig:black-box-fgm-defended},~\ref{fig:black-box-jsma-defended} and~\ref{fig:black-box-adadelta-defended}, we use a \emph{defended} surrogate model trained with defensive distillation. Figures~\ref{fig:black-box-fgm-undefended},~\ref{fig:black-box-jsma-undefended} and~\ref{fig:black-box-adadelta-undefended} use an \emph{undefended} surrogate model.}
	\label{fig:black-box}
\end{figure*}

Our results, reported in Figure~\ref{fig:black-box}, are comparable with those
obtained in the white-box setting (see Figure~\ref{fig:white-box}).
Black-box adversarial examples are somewhat more likely to be detected as outliers
than their white-box counterparts. Yet, the misclassification rates of each attack 
are comparable in the white-box and black-box settings, regardless of the fact that
the model was defended or not. This comforts the design choices made to
prevent gradient masking.

\section{Discussion}
\label{sec:discussion}

\vspace*{-0.1in}

We addressed the numerical instabilities encountered by the original
defensive distillation mechanism.
In lieu of extracting uncertainty estimates from the probability vectors at high temperature, we considered
the logits produced by multiple stochastic passes through the neural network.
Future work may build on the ideas of \textbf{transferring knowledge and uncertainty}
between models in order to improve ML
performance on other tasks in adversarial environments.

The variant proposed defends models in a comparable capacity in the
face of white-box and black-box attacks. This indicates that \textbf{the defense
is less likely to suffer from gradient masking}. While its benefits 
recede as perturbations become larger,
defensive distillation is able to improve the robustness of a neural network
in a vicinity of its test data at a reasonable
price in accuracy. 

Although the method is generic and applicable to any neural net and input domain,
our evaluation remains preliminary. The MNIST dataset is
nevertheless compelling because of \textbf{the
strong transferability across architectures, which makes defending against
black-box attacks particularly hard}.

Our results stress that \textbf{it is key to defend against two types of adversarial examples---built with
infinitesimal and finite perturbations}. In the black-box setting, the latter can be more challenging
because they are more ``transferable''. However,
AdaDelta adversarial examples
transfer less across architectures at $\kappa\simeq 2$. Thus, our results on undefended black-box attacks are pessimistic in the sense that they assumed
the adversary has perfect knowledge about the defended model's architecture.

\textbf{Defensive distillation's most appealing aspect remains that
	it does not require that the defender generate adversarial examples.}
This leaves room for another line of work combining defensive distillation
with other defenses. For example, adversarial training could be combined
with defensive distillation to learn a mapping between
adversarial examples and the correct or outlier classes. Given that adversarial training
is typically effective against finite perturbations (e.g., the FGM), it is likely
to complement defensive distillation. 

\section*{Acknowledgments}
\vspace*{0.1in}

\noindent Nicolas Papernot is supported by a Google PhD Fellowship in Security.
We thank NVIDIA for the donation of a Titan X Pascal.
Research was supported in part by the Army Research Laboratory, under
Cooperative Agreement Number W911NF-13-2-0045 (ARL Cyber Security CRA), and the
Army Research Office under grant W911NF-13-1-0421.



\end{document}